\pdfoutput=1 %
\documentclass{article}

\usepackage{xcolor}
\usepackage[preprint]{corl_2026} %

\usepackage{graphicx}
\usepackage{booktabs}
\usepackage{wrapfig}
\usepackage{xspace}
\usepackage{pifont}
\usepackage{subcaption}
\usepackage{enumitem}
\usepackage{listings}
\usepackage{multirow}
\usepackage{array}
\usepackage{placeins}
\usepackage{tikz}
\usetikzlibrary{positioning,arrows.meta,calc,fit,backgrounds,shapes.geometric}
\newcommand{\algname}{RoboSimStudio\xspace}

\newcommand{\rss}{RoboSimStudio\xspace}

\newif\ifshowtodos
\showtodosfalse

\newif\ifshowadded
\showaddedfalse
\definecolor{addedcolor}{RGB}{0,72,186}

\DeclareCaptionFont{addedfont}{\ifshowadded\color{addedcolor}\fi}

\newcommand{\cmark}{\ding{51}}

\newcommand{\pmark}{$\sim$}

\newlength{\taskpw}
\newcommand{\taskpanel}[2]{%
  \begin{tikzpicture}[inner sep=0pt, baseline=(p.south)]
    \node[anchor=south west] (p) at (0,0) {\includegraphics[width=\taskpw]{figures/tasks/#1.jpg}};
    \node[overlay, anchor=north west, fill=white, fill opacity=0.85, text opacity=1, rounded corners=1pt,
          inner sep=1.1pt, font=\fontsize{5.5}{6}\selectfont] at ([shift={(1.2pt,-1.2pt)}]p.north west) {#2};
  \end{tikzpicture}}
\newcommand{\tasksep}[1]{%
  \hspace{1.8pt}\smash{\textcolor{black!25}{\rule[-#1]{0.4pt}{\dimexpr0.75\taskpw+#1\relax}}}\hspace{1.8pt}}

\tikzset{tilechip/.style={overlay, fill=white, fill opacity=0.85, text opacity=1, rounded corners=1pt,
  inner sep=1.1pt, font=\fontsize{5.5}{6}\selectfont}}
\newcommand{\cornertile}[4]{\makebox[#1][c]{%
  \begin{tikzpicture}[inner sep=0pt, baseline=(p.south)]
    \node[anchor=south west] (p) at (0,0) {\includegraphics[width=#1]{#2}};
    \draw[black!25, line width=0.3pt] (p.south west) rectangle (p.north east);
    \if\relax\detokenize{#3}\relax\else
      \node[tilechip, anchor=north west] at ([shift={(1.2pt,-1.2pt)}]p.north west) {#3};\fi
    \if\relax\detokenize{#4}\relax\else
      \node[tilechip, anchor=south west] at ([shift={(1.2pt,1.2pt)}]p.south west) {#4};\fi
  \end{tikzpicture}}}
\newcommand{\chiptile}[3]{\cornertile{#1}{#2}{#3}{}}

\definecolor{mjviolet}{RGB}{91,63,166}
\definecolor{sdteal}{RGB}{22,160,132}
\definecolor{vbdamber}{RGB}{214,128,8}
\newcommand{\solver}[2]{{\color{#1}\rule[-0.2pt]{4.4pt}{4.4pt}}\,#2}
\newcommand{\plaintile}[2]{\cornertile{#1}{#2}{}{}}
\newlength{\mppw}
\newcommand{\mpcolumn}[3]{%
  \begin{minipage}[t]{\mppw}\centering
    \chiptile{\mppw}{figures/multiphysics/#1_film.jpg}{\textbf{#2}}\\[1.5pt]
    \plaintile{\mppw}{figures/multiphysics/#1_mech.jpg}\\[1.5pt]
    {\fontsize{5.6}{6.8}\selectfont #3\par}
  \end{minipage}}

\newlength{\rpw}\newlength{\rvw}\newlength{\rlw}
\newcommand{\tilerow}{\par\vspace{1.5pt}}
\newcommand{\tilestack}[1]{\begin{minipage}[b]{\linewidth}\offinterlineskip #1\end{minipage}}
\newcommand{\mprowlabel}[1]{\rotatebox{90}{\makebox[0.75\mppw][c]{\fontsize{6}{7}\selectfont\textbf{#1}}}}
\newcommand{\mprowlabels}[2]{%
  \begin{minipage}[t]{7pt}\mprowlabel{#1}\\[1.5pt]\mprowlabel{#2}\end{minipage}}

\lstdefinestyle{rss}{
  language=Python, basicstyle=\ttfamily\scriptsize, columns=fullflexible, keepspaces=true,
  keywordstyle=\color{blue!55!black}, commentstyle=\color{black!50}\itshape,
  stringstyle=\color{green!35!black}, showstringspaces=false,
  frame=single, rulecolor=\color{black!25}, framesep=3pt, xleftmargin=4pt, xrightmargin=4pt,
  aboveskip=2pt, belowskip=2pt}
\lstdefinestyle{rssout}{
  basicstyle=\ttfamily\scriptsize\color{black!65}, columns=fullflexible, keepspaces=true,
  frame=none, xleftmargin=4pt, aboveskip=1pt, belowskip=1pt}

\title{RoboSimStudio: An Interactive Toolkit for Rapid Real2Sim of Robot Manipulation Tasks}
\hypersetup{pdftitle={RoboSimStudio: An Interactive Toolkit for Rapid Real2Sim of Robot Manipulation Tasks},
  pdfauthor={Kaiyuan Chen, Shuangyu Xie, Uksang Yoo, Xinyue Zhu, Cael Magner, Alvin Zhu, Daehwa Kim, Eric Heiden, Erwin Coumans, Miles Macklin, Spencer Huang, Ken Goldberg}}

\author{
  Kaiyuan Chen$^{2*}$, Shuangyu Xie$^{2*}$, Uksang Yoo$^{2}$, \\  Xinyue Zhu$^{2}$, Cael Magner$^{2}$, Alvin Zhu$^{2}$, 
  Daehwa Kim$^{2}$, \\ Eric Heiden$^{1}$, Erwin Coumans$^{1}$, Miles Macklin$^{1}$, Spencer Huang$^{1}$, Ken Goldberg$^{2}$  \\
$^{*}$Equal contribution \\ 
  $^{1}$NVIDIA \quad $^{2}$University of California, Berkeley  \\
}

\begin{document}
\maketitle
\suppressfloats[t] %
\vspace{-20pt}
\begin{center}
    \centering
    
    \captionsetup{type=figure}
    \includegraphics[width=\linewidth]{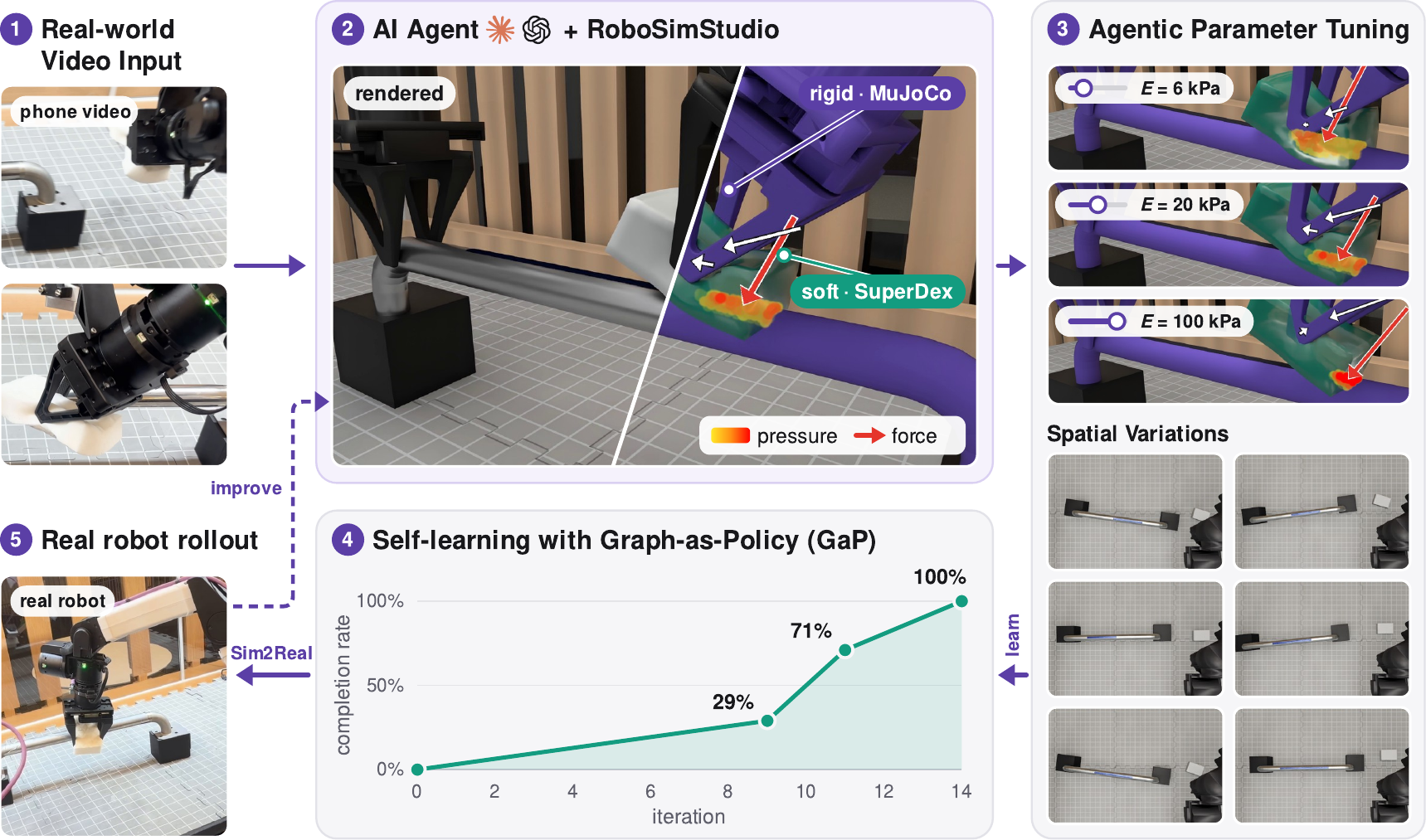}

\captionof{figure}{\textbf{RoboSimStudio} takes an input a smartphone video of
  a robot or human performing a task and outputs a multi-physics simulation that humans
  and coding agents can inspect and tune. For example, given the video of a sponge-wiping task, \rss uses Claude~\cite{claudecode2025} or Codex~\cite{openaicodex2025}  agents
  build the simulation. {\algname} couples the physics through  Newton~\cite{newton2025} and  the sponge with SuperDex~\cite{superdex2026}. A human or a coding agent can interact with the simulation and tunes its solver, contact model, and parameters in a loop, until its behavior matches the video. For robustness, the simulation is then varied in its physical parameters (e.g. Young's modulus $E$) and its spatial layout. With the varied spatial layout and physical parameters, Graph-as-Policy~\citep{gap2026} uses simulation feedback to improve success rate and cycle time before the policy runs on the real robot. }
    \label{fig:scene-title}
\end{center}%

\begin{abstract}
It has been difficult to 
accurately model the physics of the  interaction between a robot and the objects it manipulates, including contact mechanics, coupled contact dynamics, deformations. stiffness and friction.
We present {\algname} (RSS), an interactive  toolkit that makes choices of solver, contact model, and physical parameters explicit and editable, allowing both humans and coding agents to quickly create and iteratively refine simulations based on observed physical behavior.
We implement RSS with the open Newton physics simulation system and describe experiments with human-guided real2sim using (1) visual evaluation, (2) force-feedback and (3) agentic Real2Sim2Real. 
Code, video and additional information can be found on the project website:  \url{https://robosimstudio.github.io}

\end{abstract}

\section{Introduction}
Simulating robot manipulation requires modeling the interaction between a robot and the objects it manipulates, including robot kinematics, coupled contact dynamics, and deformation (Figure \ref{fig:challenge}).
Emerging simulation frameworks, such as Newton~\cite{newton2025}, Genesis~\cite{genesis2024}, and SuperDex~\cite{superdex2026}, increasingly cover these different forms of physics by supporting rigid, articulated, and deformable-body simulation within a common framework.
However, accurately modeling robot manipulation behavior to match real-world physics, or Real2Sim,  still requires choosing appropriate physics models, contact models, and physical parameters, and iteratively refining these choices against observed behavior. %
In this paper, we 
extend existing simulation frameworks to streamline Real2Sim.  

We recognize that both humans and LLM coding agents share a fundamental interaction pattern for Real2Sim:  they modify a scene, execute physical interactions, observe discrepancies, and refine the model based on feedback. Specifically, humans perform this loop through direct manipulation and immediate visual or force feedback, while coding agents~\cite{claudecode2025,openaicodex2025} use iterative reasoning and acting~\cite{yao2023react} to inspect simulator states and revise scene configurations and robot behaviors. 
Motivated by this insight, we make multi-physics simulations directly interactive and editable by both humans and coding agents to enable rapid  iterative Real2Sim refinement.

We present {\algname} (RSS), an interactive toolkit for physics-based simulation of robot manipulation tasks.
It uses Viser~\cite{yi2025viser} and Newton~\cite{newton2025} for multi-physics simulation, supports co-simulation with external physics engines, and makes choices of solver, contact model, and physical parameters explicit and editable.
An interactive  display and sliders allow a human or AI coding agent~\cite{claudecode2025,openaicodex2025} to rapidly adjust parameters to align with a real robot and environment. 
For coding agents, {\algname} exposes simulator state together with spatial-temporal text and visual feedback, allowing agents to observe the consequences of their changes and iteratively refine the simulation.

We conduct evaluation on \rss through human-guided and agentic Real2Sim experiments.
Using the interactive studio, we construct 18 contact-rich manipulation simulations involving rigid objects, articulated mechanisms, cables, cloth, and soft-body deformation.
We further demonstrate force-feedback interaction, where a human operator can control a simulated robot and perceive contact forces through a low-cost leader arm.
Finally, we conduct a pilot study of agentic Real2Sim2Real using a bimanual sponge-wiping task (Figure \ref{fig:scene-title}).
Given a smartphone video of the real task, off-the-shelf coding agents construct multi-physics simulations.
A Graph-as-Policy (GaP)~\cite{gap2026} is then improved using simulation feedback and transferred to the real robot.
We also demonstrate how a real-world failure can be used to guide an agent to revise the simulation's physical constraints.

In summary, this work makes three contributions: (1) \rss, a simulation toolkit with configurable contact models and physical parameters for contact-rich robot manipulation; (2) Interactive interfaces on \rss for humans and coding agents that streamline Real2Sim process; (3) 18 manipulation tasks with visual and force-feedback interaction, and  a pilot study on real2sim2real with \rss that coding agents construct a multi-physics simulation from a smartphone video.

\section{Related Work}
\label{sec:related}

\begin{table*}[t]
\centering
\caption{\textbf{Comparison of robotics simulation frameworks.}
Comparison of physics capabilities, interactive interfaces, and
Real2Sim workflows. \cmark~indicates built-in support,
\pmark~indicates partial support, and ``Custom'' indicates
that additional implementation is required.}
\label{tab:sim_comparison}

\renewcommand{\arraystretch}{1.2}
\setlength{\tabcolsep}{7pt}

\resizebox{\linewidth}{!}{%
\begin{tabular}{lcccccc}
\toprule
\textbf{Feature}
& \textbf{MuJoCo}
& {\textbf{IsaacLab}}
& \textbf{ManiSkill3}
& \textbf{Genesis}
& \textbf{Newton}
& \textbf{\rss} \\
\midrule

\multicolumn{7}{l}{\textbf{Physics Simulation}} \\
\addlinespace[2pt]
Rigid/articulated bodies
& \cmark & \cmark & \cmark & \cmark & \cmark & \cmark \\
Deformable objects
& \cmark & \cmark & \pmark & \cmark & \cmark & \cmark \\
Coupled multi-physics solvers
& -- & \pmark & -- & \cmark & \pmark & \cmark \\

\addlinespace[5pt]
\multicolumn{7}{l}{\textbf{Interactive Interfaces}} \\
\addlinespace[2pt]
Interactive scene inspection
& \cmark & \cmark & \cmark & \cmark & \cmark & \cmark \\
Physics parameter tuning
& \pmark & \cmark & \pmark & \pmark & \pmark & \cmark \\
Structured feedback for coding agents
& Custom & Custom & Custom & Custom & Custom & \cmark \\
Shared human-agent refinement
& Custom & Custom & Custom & Custom & Custom & \cmark \\

\addlinespace[5pt]
\multicolumn{7}{l}{\textbf{Real2Sim Workflows}} \\
\addlinespace[2pt]
Video-driven agentic Real2Sim
& -- & -- & -- & -- & -- & \cmark \\
Force-feedback Real2Sim
& -- & \pmark & -- & -- & -- & \cmark \\
Failure-driven simulation refinement
& -- & -- & -- & -- & -- & \cmark \\

\bottomrule
\end{tabular}%
}
\end{table*}

\paragraph{Physics Simulation for Robot Learning.}
Physics simulation has enabled large-scale robot learning by providing environments for policy training and evaluation.
MuJoCo~\citep{todorov2012mujoco} supports articulated robot dynamics and contact-rich manipulation, while GPU-accelerated frameworks such as Isaac Gym and Isaac Lab~\citep{makoviychuk2021isaacgym,mittal2023orbit,isaaclab2025}, ManiSkill3~\citep{tao2025maniskill3}, MuJoCo Playground~\citep{zakka2025playground}, and Genesis~\citep{genesis2024} support parallel simulation for robot learning.
Frameworks such as robosuite~\citep{zhu2020robosuite} further provide standardized manipulation environments.
More recently, Newton~\citep{newton2025}, built on Warp~\citep{warp2022}, integrates multiple physics solvers, including MuJoCo-Warp~\citep{mujocowarp2025}, Vertex Block Descent (VBD)~\citep{chen2024vbd}, and XPBD~\citep{macklin2016xpbd}.
Other physics engines, such as SuperDex~\citep{superdex2026}, provide specialized models for contact-rich soft-body simulation.
While these frameworks provide increasingly diverse physics capabilities, constructing manipulation environments still requires selecting appropriate solvers, configuring contact models, and tuning physical parameters.
\rss builds on Newton and supports co-simulation with external physics engines, allowing different physics models to be composed within the same scene and interactively refined by humans and coding agents.

\paragraph{Agentic Robotics.}
Large language models (LLMs) are now being used in robotics for task planning, robot control, and policy generation.
Early approaches use language models to ground instructions in robot affordances~\citep{ahn2022saycan}, generate plans and executable robot programs~\citep{liang2023cap,singh2023progprompt,huang2023voxposer,vemprala2024chatgpt}, and improve robot behavior through feedback~\citep{huang2022innermonologue}.
More recent works use LLMs to generate reward functions~\citep{yu2023l2r,ma2024eureka}, visual question answering~\cite{chen2025robo2vlm, xie2026robovista}, domain randomization strategies~\citep{ma2024dreureka}, and simulation tasks for robot learning~\citep{wang2024gensim,hua2024gensim2,ye2025video2policy}.
With iterative reasoning and acting frameworks such as ReAct~\citep{yao2023react}, general-purpose coding agents~\citep{claudecode2025,openaicodex2025} can modify code, execute programs, and refine their solutions based on feedback.
Such agents have been applied to reconstructing articulated objects~\citep{mandi2025real2code}, robot data collection~\citep{ahn2024autort}, and iterative policy improvement, such as CaP-X~\citep{fu2026cap} and GaP~\cite{gap2026}.
{MAESTRO~}\citep{shi2025maestro} {orchestrates robotics modules through VLM-written programs, ASPIRE~}\citep{lu2026aspire} {grows a reusable code skill library across many simulated tasks, and ENPIRE~}\citep{xiao2026enpire} {lets coding agents improve policies through real-world trial and error.}
{DexAgent~}\citep{wang2026dexagent} {turns a single egocentric human video into dexterous robot trajectories, and RPG~}\citep{wang2026rpg} {practices tasks in simulation and revises skills from diagnosed failures before real-world deployment.}
A coding agent has also solved the Push-T benchmark without demonstrations by running its own experiments in simulation~\citep{xie2026pusht}.
The effectiveness of these agents also depends on their ability to interact with external environments and access execution feedback~\citep{yang2024sweagent}.
\rss builds on this insight by providing coding agents with an interactive physics simulation environment, where scene configurations, physics models, and simulation feedback are directly accessible for agentic Real2Sim construction and refinement.

\paragraph{Real2Sim.}
Real2Sim approaches reconstruct simulation environments from real-world observations to support robot policy training and evaluation. Prior work has explored constructing digital twins and digital cousins from images, videos, and other observations~\citep{torne2024rialto,chen2024urdformer,dai2024acdc,le2024articulate,li2024simpler,qureshi2024splatsim}. Other approaches use generative models and LLMs to automatically construct simulation tasks, environments, and reward functions~\citep{wang2024gensim,wang2024robogen,yang2024holodeck,ma2024eureka}, or generate robot policies directly~\citep{liang2023cap,gap2026}. Blox-Net~\citep{goldberg2025bloxnet} uses vision-language models to design block assemblies, refines the designs through perturbation analysis in simulation, and builds them with a real robot. These approaches reduce the effort required for scene construction and policy development, but often focus on specific reconstruction or generation pipelines. Recent frontier models, such as Claude Fable 5.1~\citep{claudecode2025}, GPT-6 Astra~\citep{openaicodex2025}, and Gemini 4 Argon~\citep{google2026argon}, can assist in reconstructing visually plausible 3D scenes from videos, including producing Blender~\citep{blender2026} renderings. However, visual realism does not guarantee physical accuracy, particularly for contact-rich interactions and deformable objects such as cables, cloth, and sponges. RoboSimStudio therefore exposes editable physics models, contact parameters, and simulator feedback, enabling humans and coding agents to iteratively refine simulated physical behavior to match real-world observations rather than appearance alone.

\section{RoboSimStudio}
\label{sec:system}

\begin{figure}[t]
  \centering
  \includegraphics[width=\textwidth]{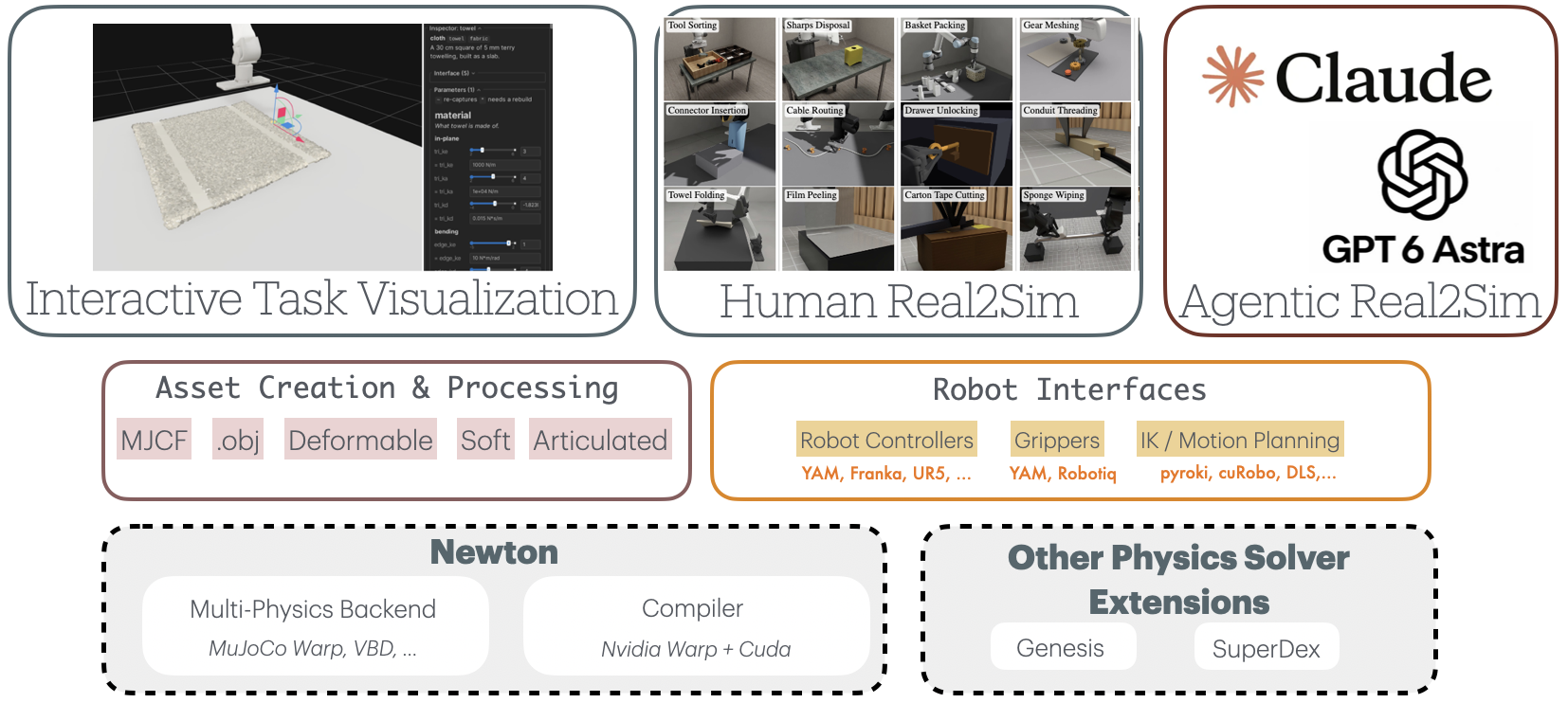}
\caption{\textbf{\rss Architecture.}
\rss integrates asset creation, robot interfaces, and multi-physics simulation through Newton and external physics solvers.
It supports interactive task visualization, human-guided Real2Sim, and agentic Real2Sim through a shared simulation framework.}
  \label{fig:architecture}
\end{figure}

Figure~\ref{fig:architecture} shows the architecture of \rss.
Physics is mainly provided by Newton~\citep{newton2025} and its solvers, with external physics engines that can be co-simulated by exchanging contact and forces.
\rss composes a scene with robots, objects, and physics into one simulation while keeping modeling choices, including the solver, contact model, and physical parameters, explicit and editable.
The composed scene can be accessed through code, an interactive studio for a human, or an iterative interface for a coding agent to support task creation and agentic real2sim.

\textbf{Robot Interfaces.}
\rss provides a common robot interface for Franka Panda, UR5e, and single- and bimanual YAM, and can be extended to additional robots.
A robot is specified declaratively by its model, joints, mounting, and hand geometry, and exposes common interfaces for forward and inverse kinematics, motion generation, and control.
Inverse kinematics supports interchangeable solvers, such as pyroki~\citep{kim2025pyroki}, while motions include joint-space and Cartesian moves, guarded contact motions, and collision-aware planning, such as cuRobo~\citep{sundaralingam2023curobo}.
Policies can command the robot in joint or end-effector action spaces, while privileged simulator state about manipulated objects is kept separate from observations available to the robot.

\begin{figure}[t]
  \centering
  \setlength{\mppw}{\dimexpr(\textwidth-7pt-20pt)/5\relax}%
  \mprowlabels{Task}{Multi-physics}\hfill
  \mpcolumn{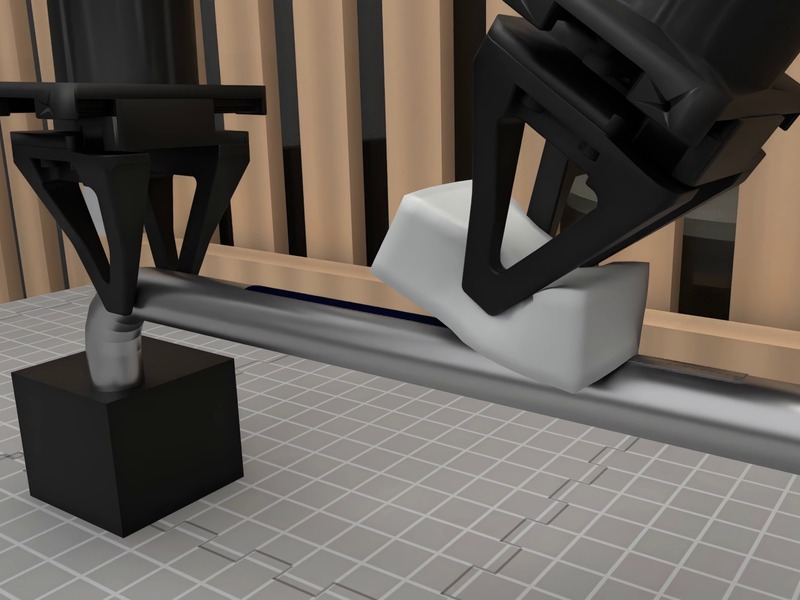}{Sponge Wiping}{\solver{mjviolet}{MuJoCo-Warp}\enspace\solver{sdteal}{FEM}}\hfill
  \mpcolumn{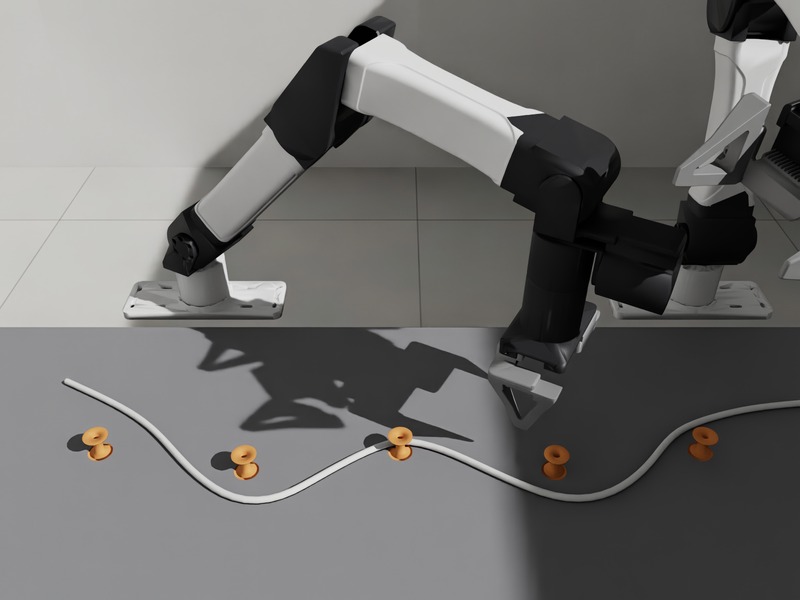}{Cable Routing}{\solver{mjviolet}{MuJoCo-Warp}\enspace\solver{vbdamber}{VBD}}\hfill
  \mpcolumn{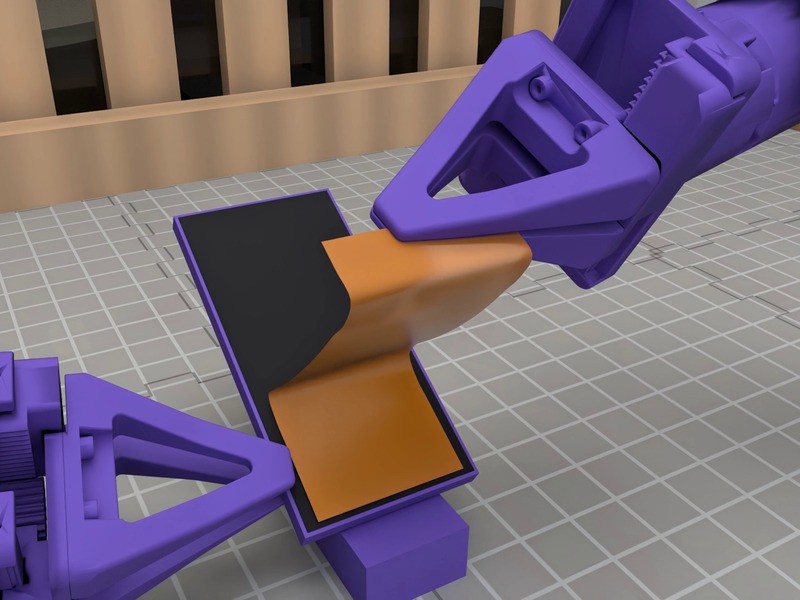}{Film Peeling}{\solver{mjviolet}{MuJoCo-Warp}\enspace\solver{vbdamber}{VBD}}\hfill
  \mpcolumn{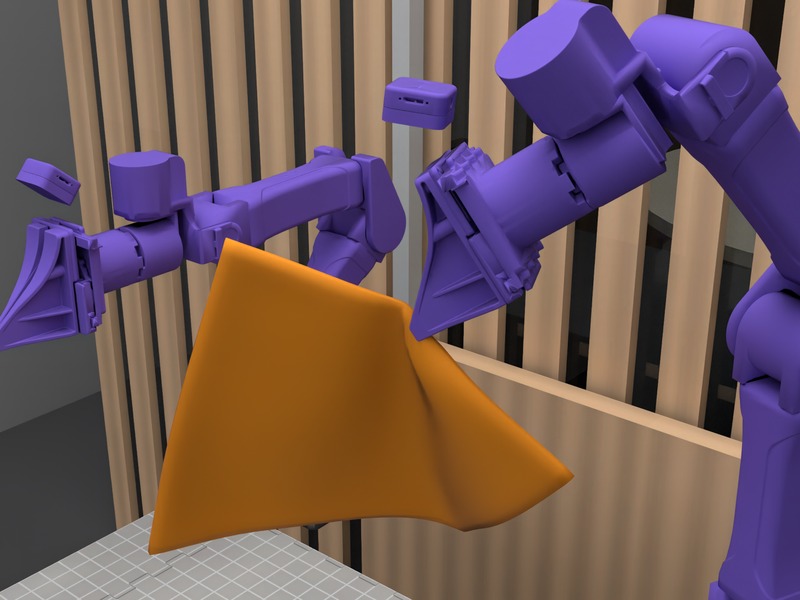}{Towel Handling}{\solver{mjviolet}{MuJoCo-Warp}\enspace\solver{vbdamber}{VBD}}\hfill
  \mpcolumn{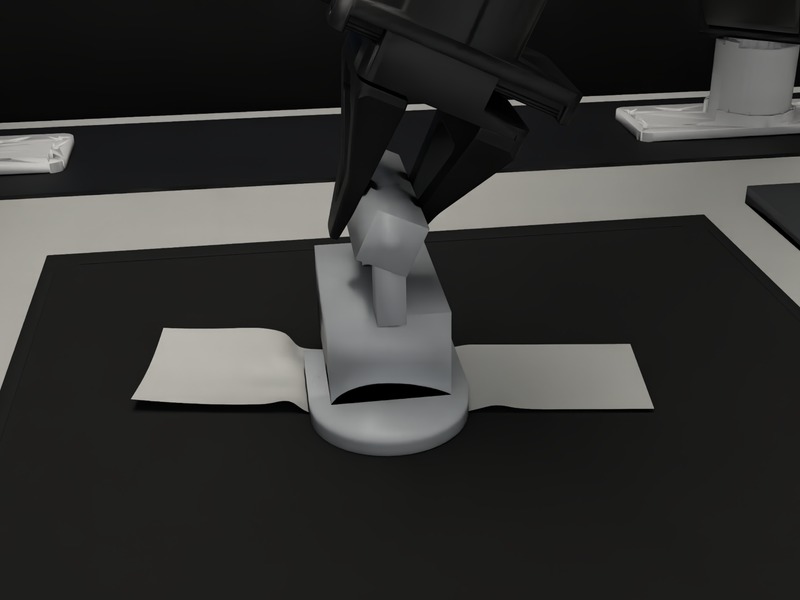}{Heat-Transfer Printing}{\solver{mjviolet}{MuJoCo-Warp}\enspace\solver{vbdamber}{VBD}}
\caption{\textbf{Multi-physics coupling in \rss.}
Five manipulation tasks using different physics solvers and
coupling methods. Top: rendered simulation scenes.
Bottom: visualization of the physics solvers used for each
object, including MuJoCo-Warp (violet), finite-element
simulation (FEM, teal), and Vertex Block Descent (VBD,
amber). All tasks use the same robot interface and scene
description framework in \rss.}
  \label{fig:multiphysics}
\end{figure}

\textbf{Multi-Physics Simulation and Coupling.}
Manipulation scenes often combine articulated robots, rigid objects, and deformable materials, for which no single physical model is sufficient.
In \rss, objects specify what they are, while the scene specifies which solver models each type of physics.
By default, articulated robots use MuJoCo-Warp~\citep{mujocowarp2025}, cables and cloth use Vertex Block Descent~\citep{chen2024vbd}, and soft bodies can be simulated by external engines~\citep{superdex2026,genesis2024} co-simulated with Newton.
\rss couples these solvers by exchanging contact forces between Newton solvers or poses and contact wrenches with external engines.
The solver, coupling method, contact model, and material parameters remain explicit entries in the scene and can therefore be changed without rewriting the task.
Figure~\ref{fig:multiphysics} shows five manipulation tasks
that use different physics solvers and coupling methods.

\begin{figure}[t]
  \centering
  \includegraphics[width=\linewidth]{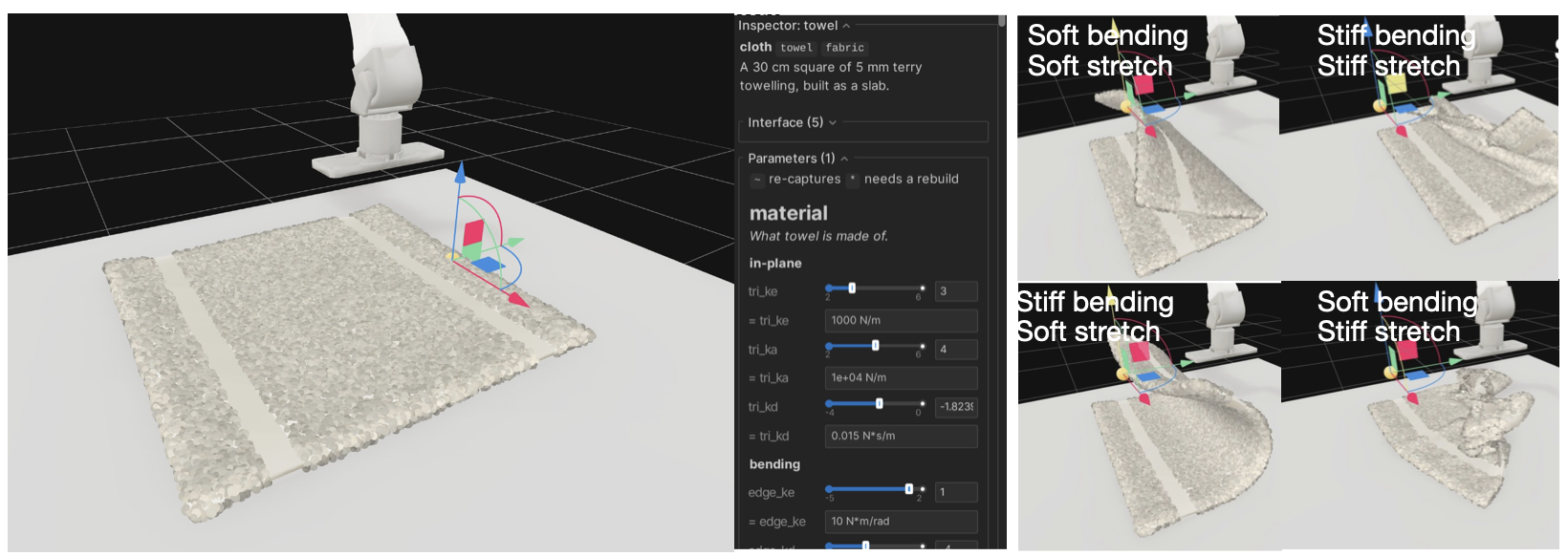}
  \caption{\textbf{\rss Interface.} Left: a towel and its inspector with material controls, for example, tri\_ke controls the stretch stiffness and edge\_ke controls the bending stiffness. Right: one towel, four bending and stretch stiffness: the same dragged fold at the same instant.}
  \label{fig:human}
\end{figure}

\textbf{Interactive Interface for Humans.}
Humans interact with \rss through a browser-based studio built on Viser~\citep{yi2025viser}.
Objects can be selected, dragged, and inspected, and robot arms can be controlled directly, allowing a user to probe physical behavior before writing a policy.
Each object's inspector exposes its physical parameters with their units and ranges.
Parameters such as stiffness and damping can be changed while the simulation is running, while parameters that require rebuilding the scene are marked explicitly.
Figure~\ref{fig:human} shows the same towel fold under four choices of bending and stretch stiffness, illustrating how simulation tuning becomes direct interaction with the resulting physical behavior.

\textbf{Iterative Interface for Agents.}
Given as input video of a real task being performed, 
a coding agent can tune a simulation through the same basic loop: modify the scene, execute a behavior, and observe the result.
\rss organizes a rollout into robot motion primitives and converts simulator state into spatial-temporal feedback that describes target error, object motion, contact events, grasping or slipping, and progress toward the task goal.
Because this feedback is computed from simulator state, it exposes physical events such as contact that may be difficult to infer from images alone.
Together with visual observations, this allows an agent to attribute an outcome to the action that caused it, modify either the policy or the simulation parameters, and run the task iteratively until the simulation behavior matches the input video.

\textbf{Human-in-the-Loop Agentic Real2Sim.}
\rss allows humans and coding agents to interact with the same simulation during Real2Sim.
While the coding agent constructs and modifies the scene, a human can observe the simulation through the interactive studio built on Viser, directly adjust physical parameters, and provide feedback to guide the agent.
Changes made by the agent are reflected in the studio, allowing the human to evaluate simulated behavior against real-world observations.
The human can also provide instructions and feedback asynchronously without interrupting the agent's execution.
This combines human guidance on task behavior and physical realism with the agent's ability to construct scenes, debug simulation errors, and perform repeated evaluations.
For example, when tuning a cable simulation, a human can use sliders in the Viser interface to adjust stiffness, damping, and contact parameters while observing the resulting cable behavior in real time.
This provides immediate visual feedback for parameter exploration, while the coding agent can use the same parameters and simulation feedback for automated evaluation and refinement.
The shared interface allows humans to provide feedback at a higher frequency and guide the agent toward appropriate physics models and parameters without interrupting its workflow.

\section{Case Studies}

\begin{figure}[t]
  \centering
  \setlength{\taskpw}{\dimexpr(\textwidth-21pt)/6\relax}%
  \setlength{\lineskip}{0pt}%
  \taskpanel{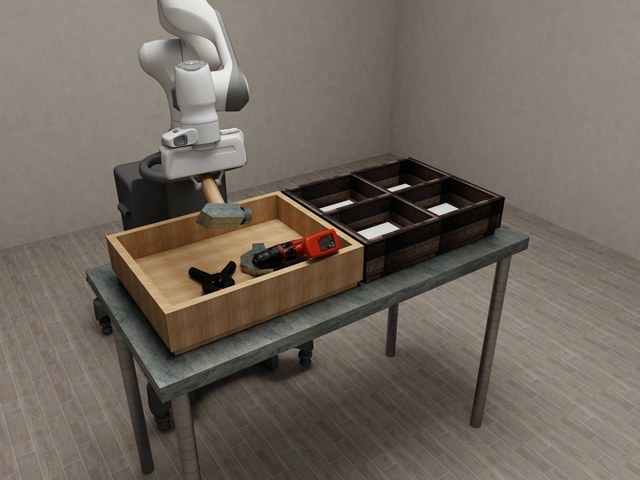}{Tool Sorting}\tasksep{0.5pt}%
  \taskpanel{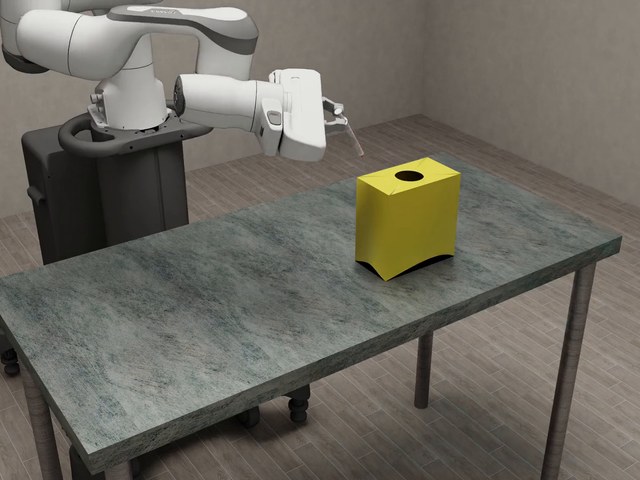}{Sharps Disposal}\tasksep{0.5pt}%
  \taskpanel{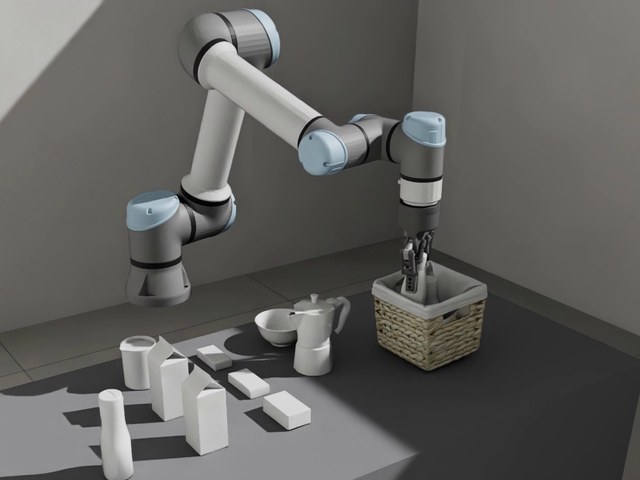}{Basket Packing}\tasksep{0.5pt}%
  \taskpanel{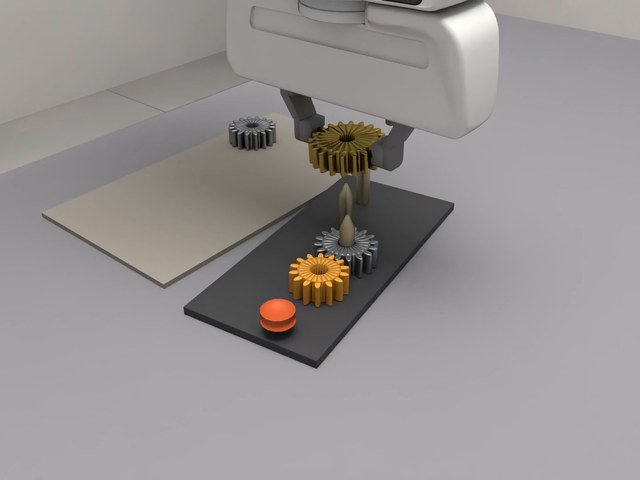}{Gear Meshing}\tasksep{0.5pt}%
  \taskpanel{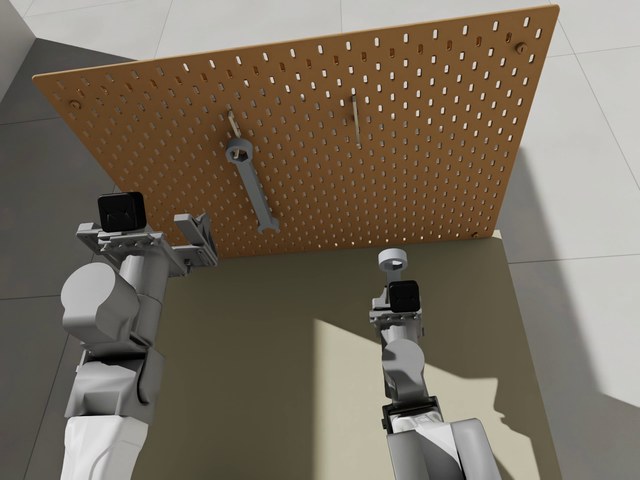}{Tool Hanging}\tasksep{0.5pt}%
  \taskpanel{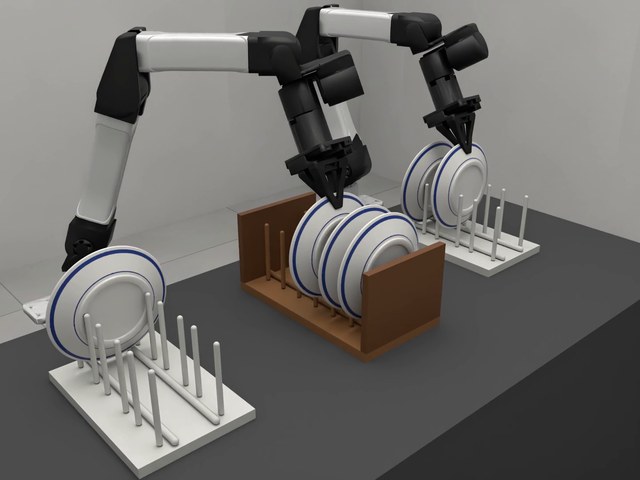}{Plate Racking}\\[0.5pt]
  \taskpanel{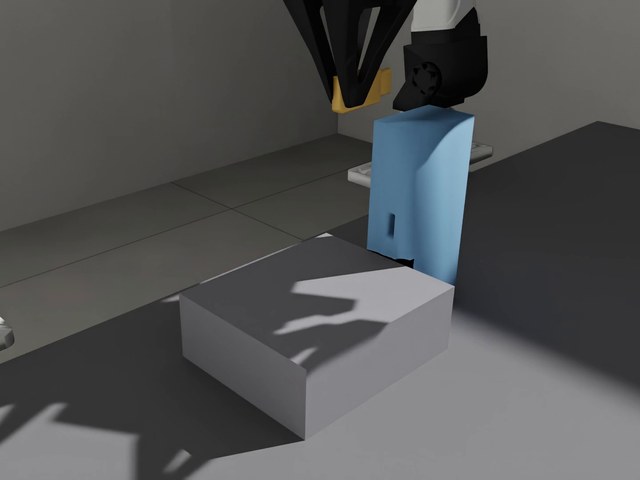}{Connector Insertion}\tasksep{0.5pt}%
  \taskpanel{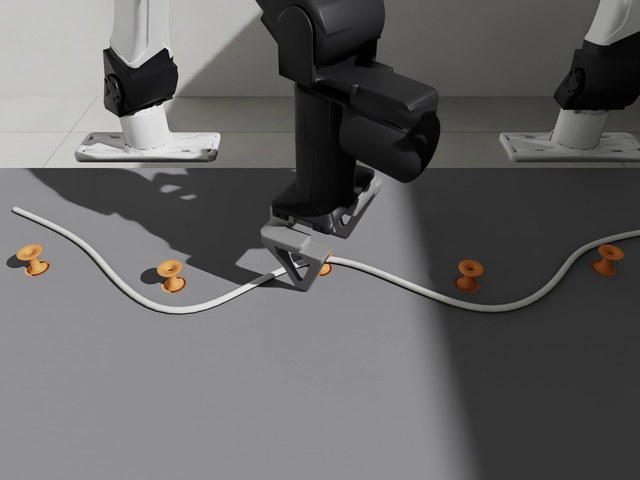}{Cable Routing}\tasksep{0.5pt}%
  \taskpanel{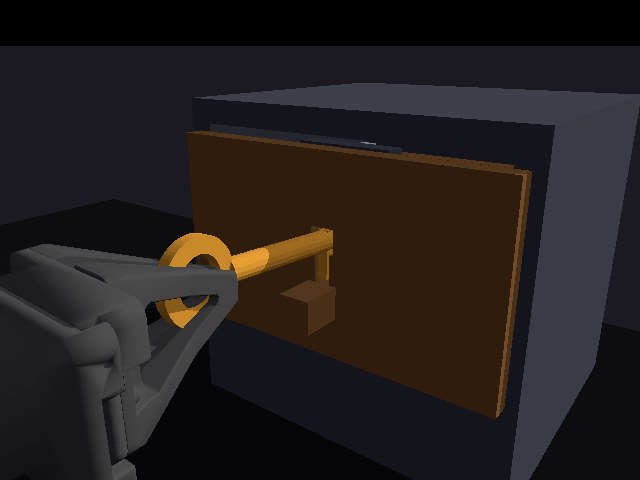}{Drawer Unlocking}\tasksep{0.5pt}%
  \taskpanel{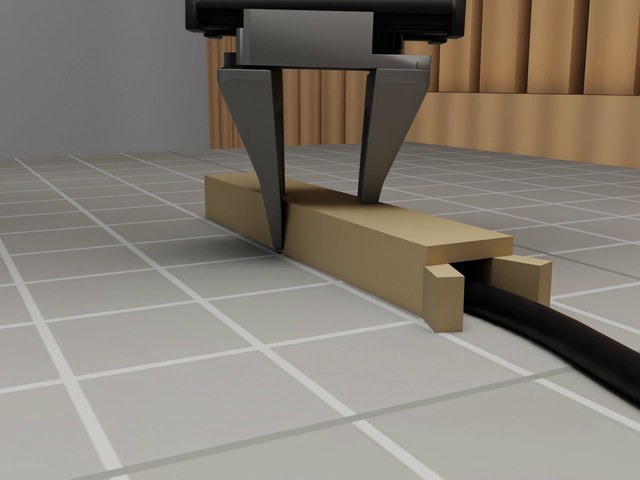}{Conduit Threading}\tasksep{0.5pt}%
  \taskpanel{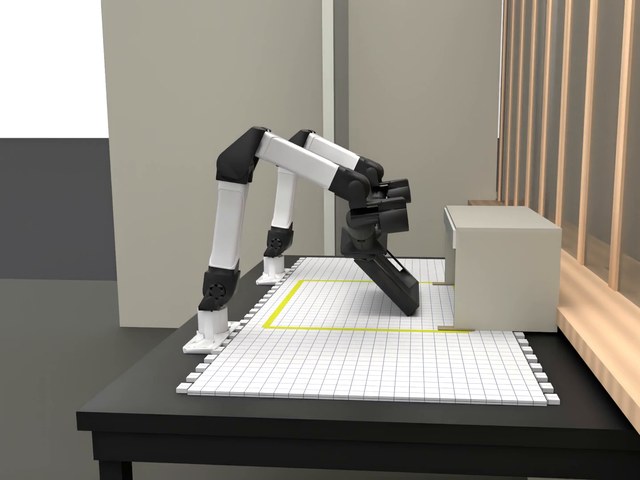}{Crate Flipping}\tasksep{0.5pt}%
  \taskpanel{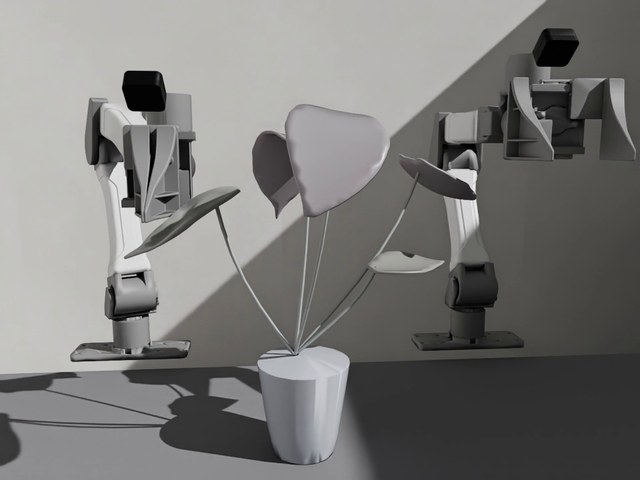}{Leaf Pressing}\\[0.5pt]
  \taskpanel{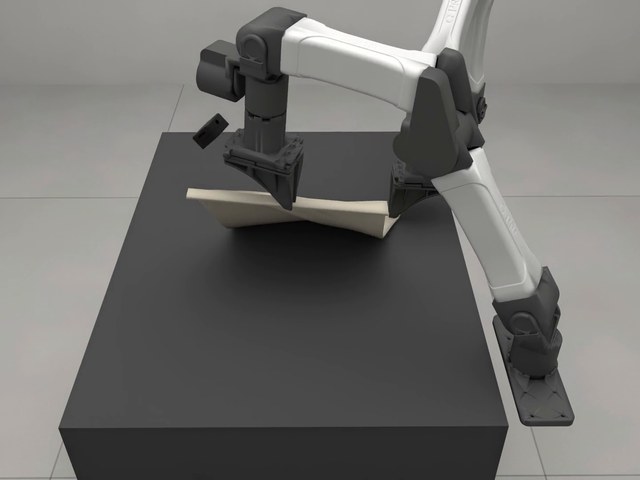}{Towel Folding}\tasksep{0pt}%
  \taskpanel{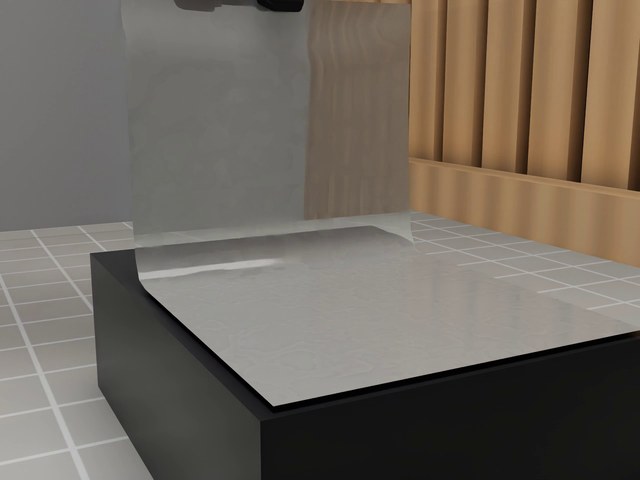}{Film Peeling}\tasksep{0pt}%
  \taskpanel{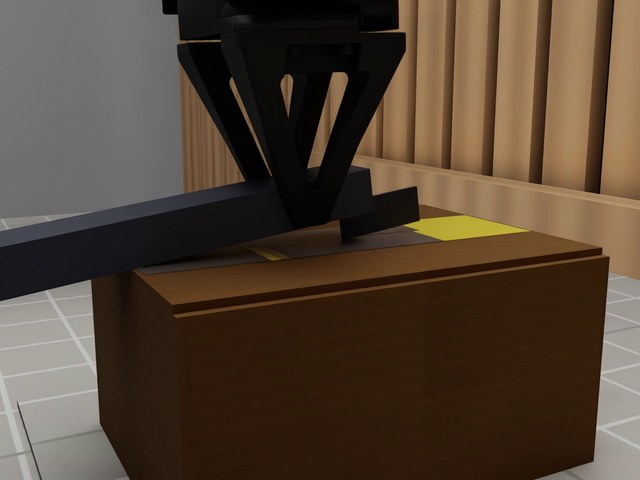}{Carton Tape Cutting}\tasksep{0pt}%
  \taskpanel{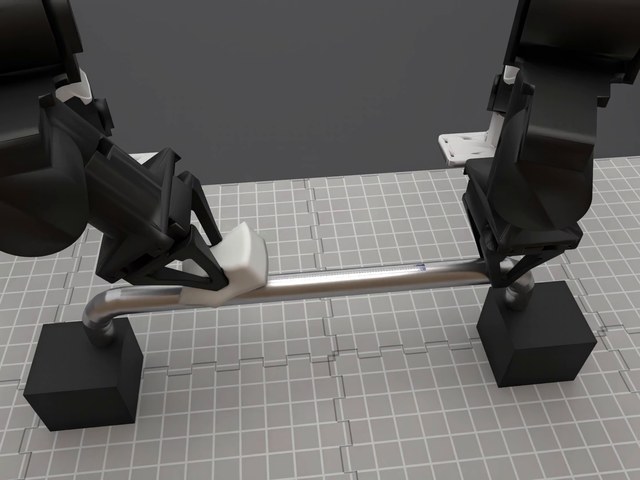}{Sponge Wiping}\tasksep{0pt}%
  \taskpanel{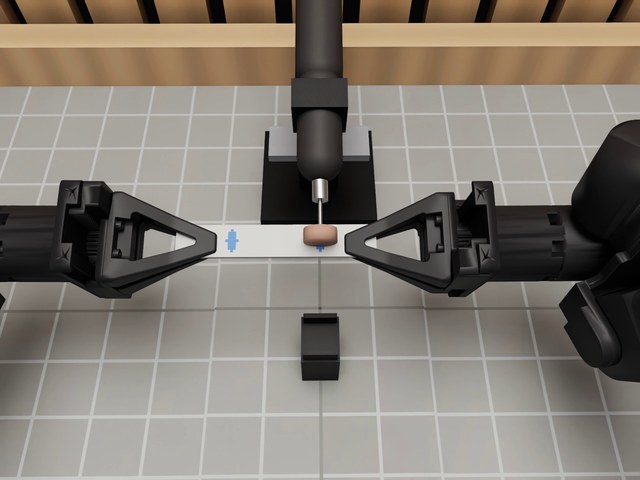}{Strip Polishing}\tasksep{0pt}%
  \taskpanel{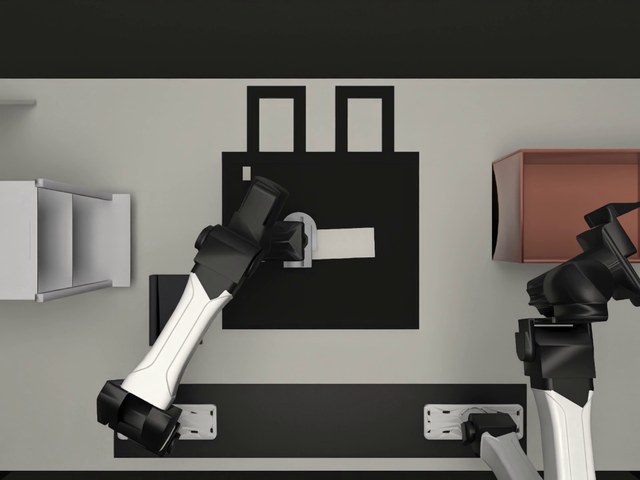}{Heat-Transfer Printing}%
  \caption{\textbf{18 Challenging contact-rich deformable manipulation task simulations created by \rss} }
  \vspace{-10pt}
  \label{fig:tasks}
\end{figure}

\subsection{Human Real2Sim with Visual Feedback}

We first study human-guided Real2Sim using visual feedback.
Starting from observations of a real manipulation task, a user constructs the corresponding scene in \rss by selecting robot and object assets, arranging their geometry, and choosing the relevant physics models.
The interactive studio then supports a tight visual refinement loop: the user executes the task, compares the simulated motion and deformation with the real behavior, and directly adjusts contact and material parameters until the two qualitatively agree.
Using this workflow, we created the 18 manipulation simulations shown in Figure~\ref{fig:tasks}, spanning rigid-body contact, articulated motion, cables, cloth, and soft-body deformation.
For example, three of these tasks highlight different challenges in simulating contact-rich manipulation:
We highlight five examples that require different physics models and contact handling:
\begin{itemize}[leftmargin=*, nosep]
    \item \emph{Connector insertion.} Simulating precise assembly requires accurate geometric contact between objects with very small clearances. The task involves inserting a USB-C plug into its receptacle with sub-millimeter clearance. \rss supports mesh-based contact and adjustable collision resolution to capture the geometry required for insertion.

    \item \emph{Leaf pressing.} Simulating thin deformable objects requires capturing their bending behavior and recovery under contact. The task involves pressing a curved leaf and releasing it to recover its shape. \rss supports shell-based simulation with adjustable solver and contact parameters to model the leaf's deformation.

    \item \emph{Strip polishing.} Simulating surface processing requires modeling physical changes caused by contact forces and relative motion. The task involves polishing a plastic strip against a rotating tool to remove surface marks. \rss exposes contact forces from the physics solver, allowing material removal to be modeled using Preston's equation~\cite{preston1927theory}.
\end{itemize}

\subsection{Human Real2Sim with Force Feedback}
While visual feedback allows users to inspect simulated behavior, it can be difficult to assess contact interactions through observation alone. For example, to perform tool insertion, it would be better for the user to feel the contact and friction without visually checking the contact between geometries.  
To complement visual inspection, \rss supports force-feedback teleoperation, allowing operators to physically perceive simulated contact and evaluate the underlying contact models and physical parameters.
\rss uses a low-cost Gello leader arm~\citep{wu2024gello} to drive a simulated robot in \rss, while contact in simulation is rendered back to the operator through the leader's motors.
The leader's joint angles command the simulated arm, and differences between the commanded and simulated configurations provide a feedback signal when contact prevents the simulated robot from following the operator.
To 
\begin{wrapfigure}{r}{0.3\linewidth}
    \centering
    \includegraphics[width=\linewidth]{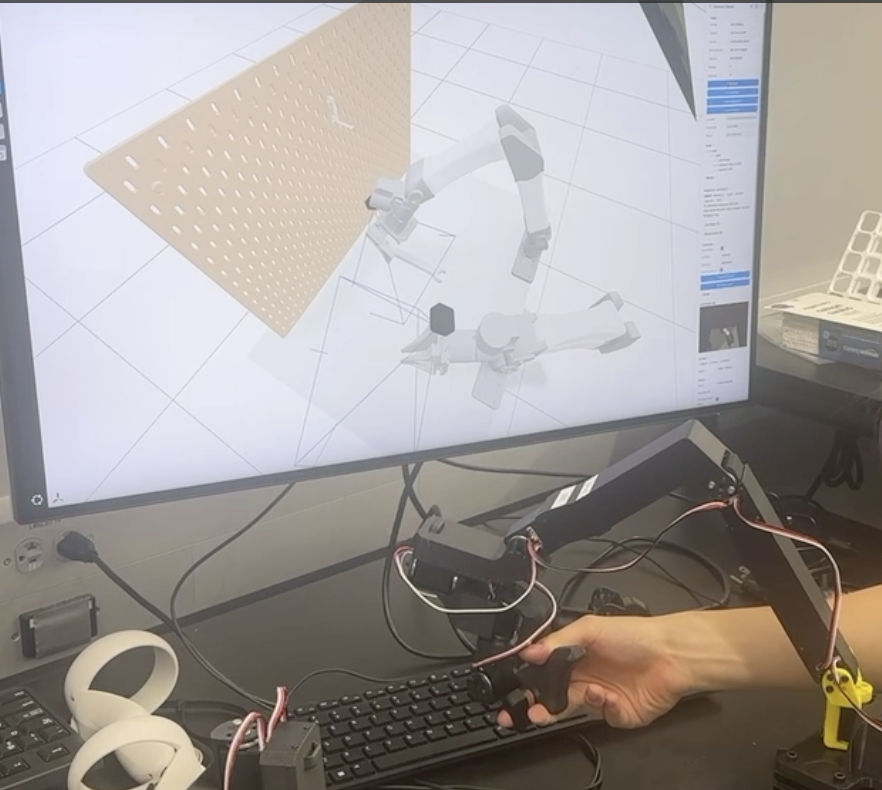}
    \caption{Tool-hanging Teleoperation with \rss}
    \label{fig:teleop}
    \vspace{-10pt}
\end{wrapfigure}
avoid rendering ordinary simulation lag as contact, \rss uses a speed-dependent deadband and releases feedback whenever the simulation pauses or falls behind.
The gripper similarly uses its measured, rather than commanded, opening so that grasping an object can be distinguished 
from closing in free space.
The human operator therefore feels the forces produced by the chosen solver and physical parameters, making teleoperation useful both for collecting demonstrations and for directly evaluating the simulated contact model.
In the tool-hanging task shown in Figure~\ref{fig:teleop}, the operator must pick up a thin tool from the table and insert it into a hanger.
Because the critical interaction occurs at contact between the tool and the hanger, force feedback from \rss helps the operator perceive and adjust the motion accordingly.

\begin{figure}[t]
  \centering
  \setlength{\rpw}{\dimexpr(\linewidth-8pt)/3\relax}%
  \setlength{\rvw}{\dimexpr(\rpw-3pt)/3\relax}%
  \setlength{\rlw}{\dimexpr(\rpw-1.5pt)/2\relax}%

  \begin{subfigure}[t]{\rpw}
    \includegraphics[width=\linewidth]{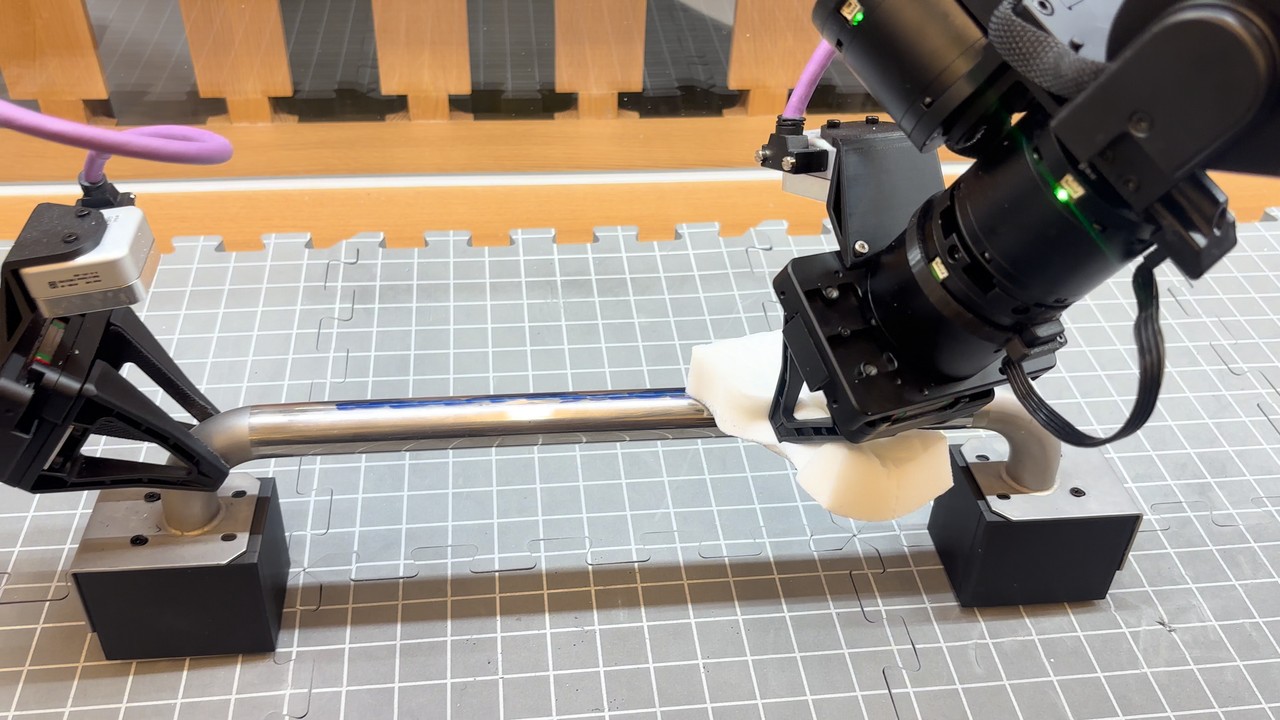}
    \caption{Real-world input video.}
    \label{fig:r2s2r:input}
  \end{subfigure}\hfill
  \begin{subfigure}[t]{\rpw}
    \includegraphics[width=\linewidth]{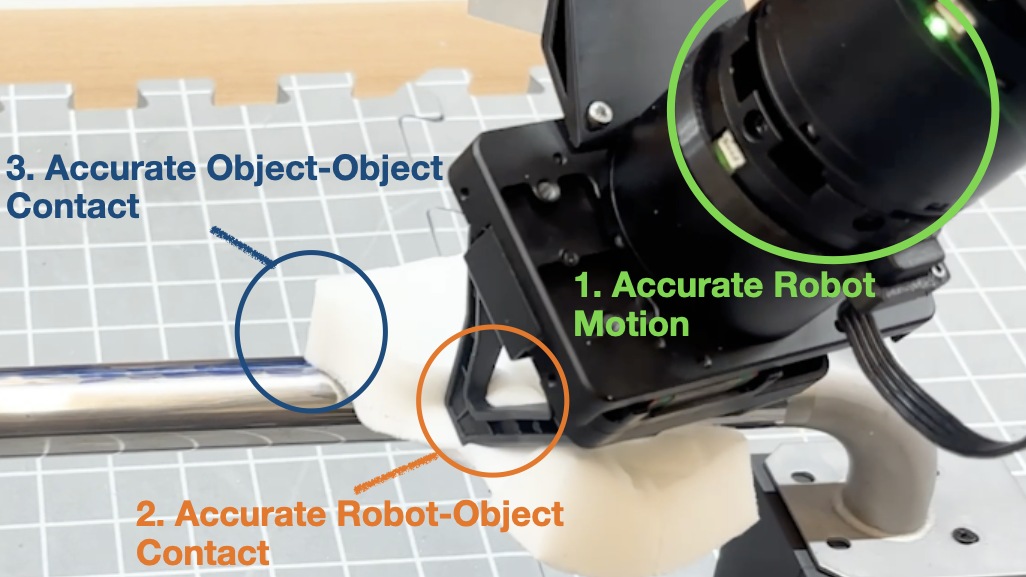}
    \caption{Task challenges.}
    \label{fig:challenge}
  \end{subfigure}\hfill
  \begin{subfigure}[t]{\rpw}
    \tilestack{%
      \chiptile{\rlw}{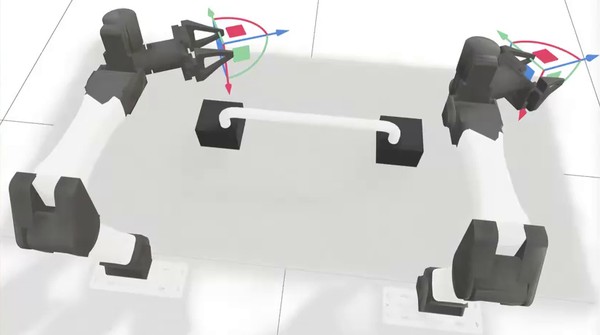}{14 min: arms, bar}\hfill
      \chiptile{\rlw}{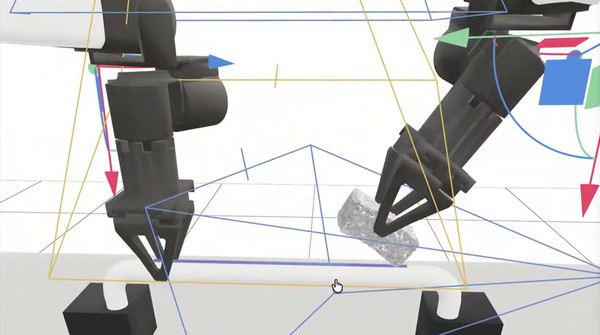}{42 min: first wipe}\tilerow
      \chiptile{\rlw}{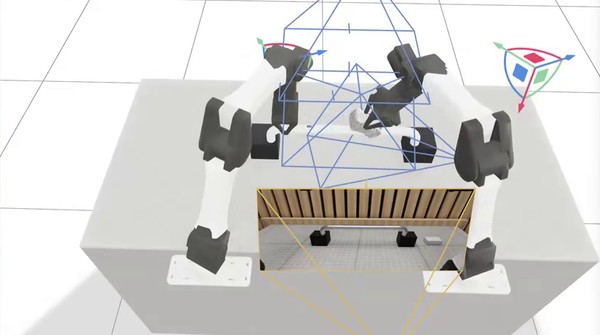}{51 min: scenery}\hfill
      \chiptile{\rlw}{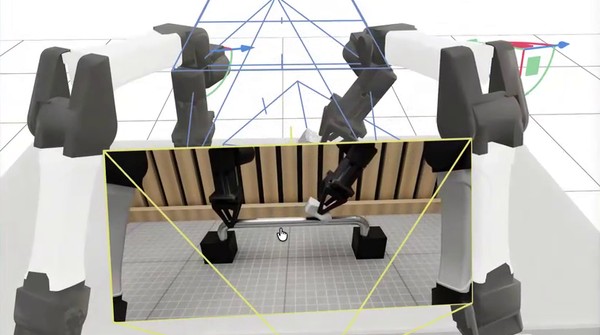}{56 min: verified}}
    \caption{Agentic Real2Sim construction.}
    \label{fig:r2s2r:build}
  \end{subfigure}\\[4pt]

  \begin{subfigure}[t]{\rpw}
    \includegraphics[width=\linewidth]{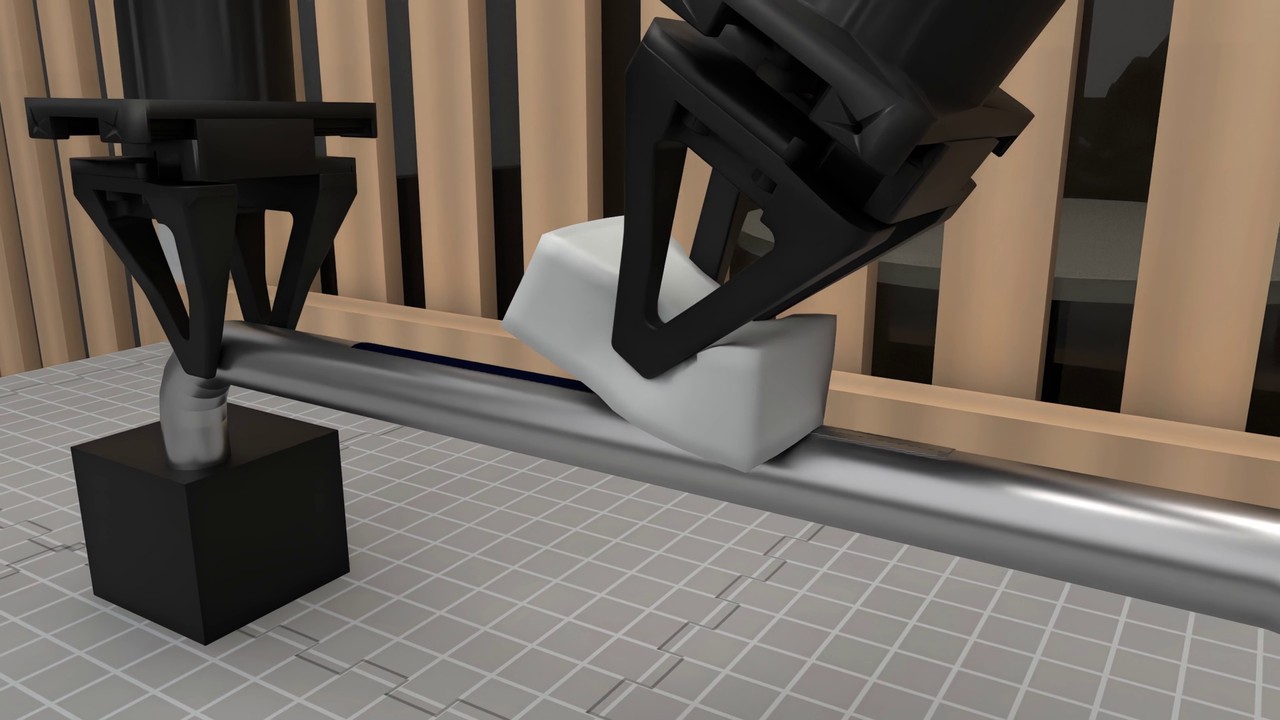}
    \caption{Multi-physics simulation in \rss.}
    \label{fig:r2s2r:twin}
  \end{subfigure}\hfill
  \begin{subfigure}[t]{\rpw}
    \tilestack{%
      \cornertile{\rvw}{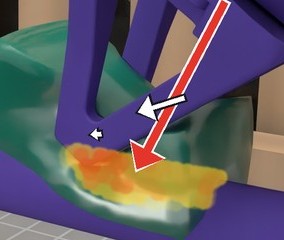}{physical variation}{6\,kPa}\hfill
      \cornertile{\rvw}{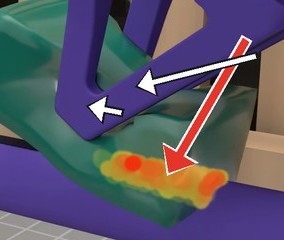}{}{20\,kPa}\hfill
      \cornertile{\rvw}{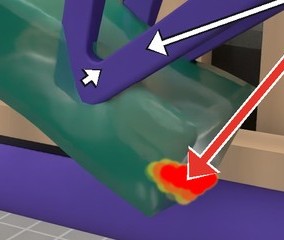}{}{100\,kPa}\tilerow
      \chiptile{\rvw}{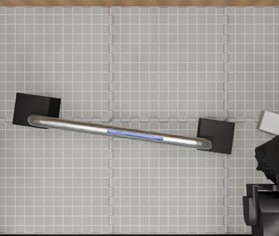}{spatial variation}\hfill
      \plaintile{\rvw}{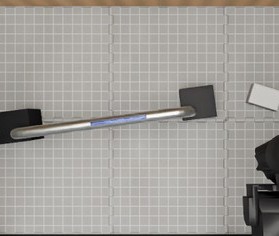}\hfill
      \plaintile{\rvw}{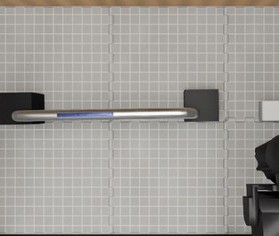}}
    \caption{Physical and spatial variations.}
    \label{fig:r2s2r:var}
  \end{subfigure}\hfill
  \begin{subfigure}[t]{\rpw}
    \tilestack{%
      \chiptile{\rlw}{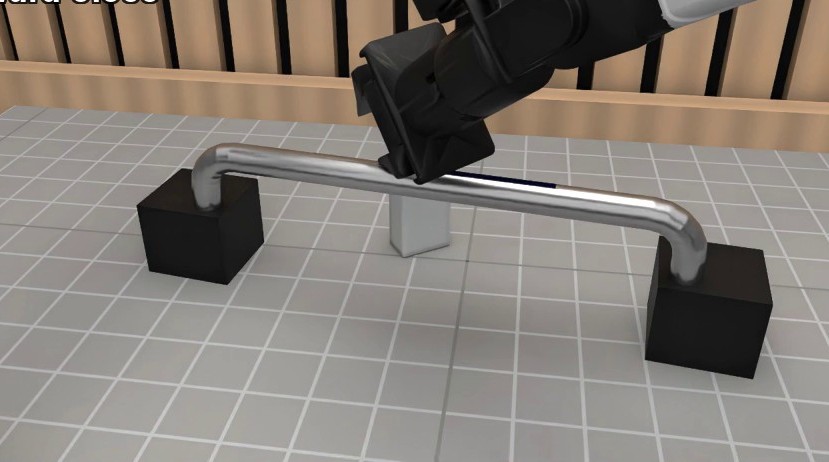}{iter.\ 0: 0\%}\hfill
      \chiptile{\rlw}{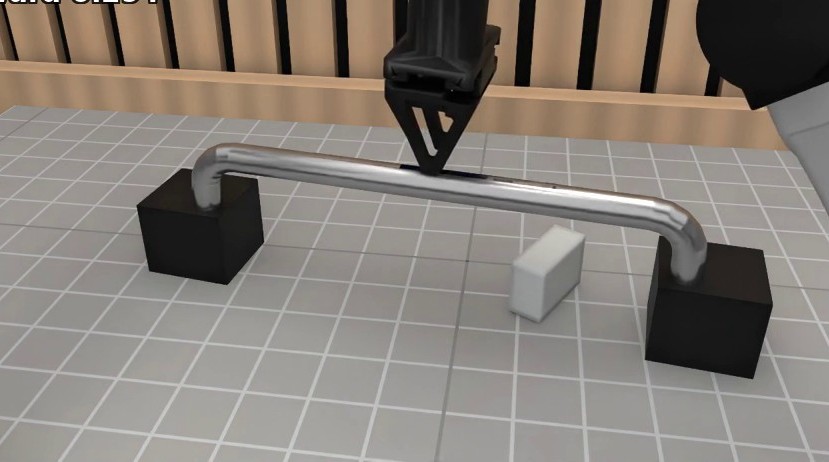}{iter.\ 9: 29\%}\tilerow
      \chiptile{\rlw}{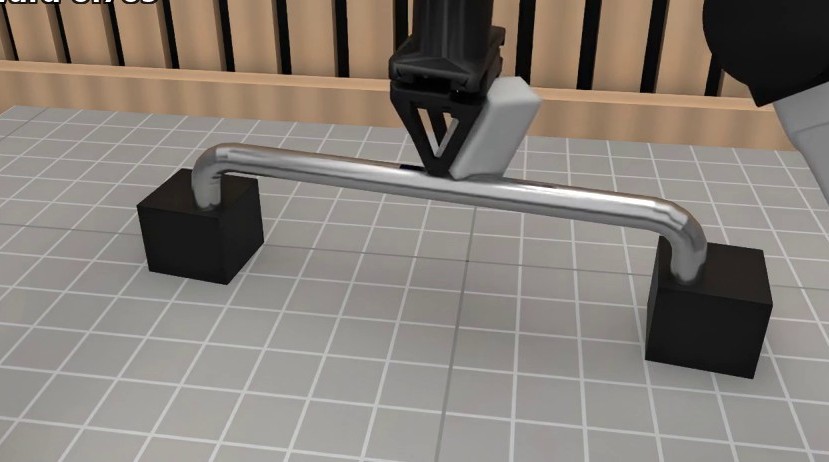}{iter.\ 11: 71\%}\hfill
      \chiptile{\rlw}{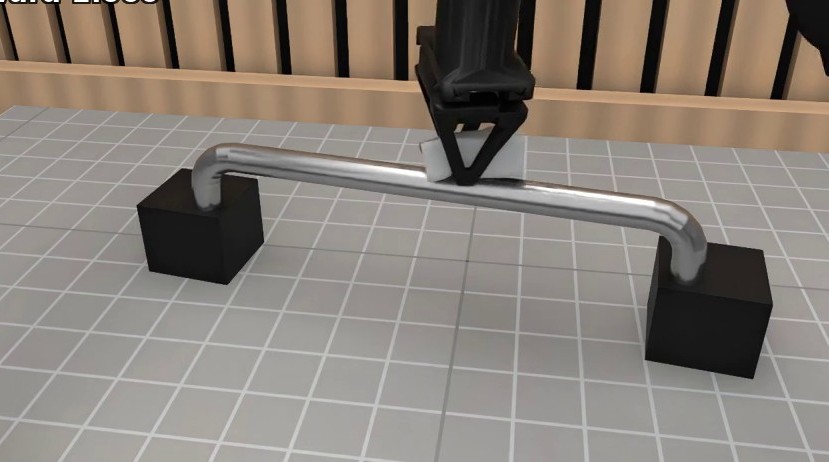}{iter.\ 14: 100\%}}
    \caption{Policy improvement with GaP.}
    \label{fig:r2s2r:learn}
  \end{subfigure}\\[4pt]

  \begin{subfigure}[t]{\rpw}
    \includegraphics[width=\linewidth]{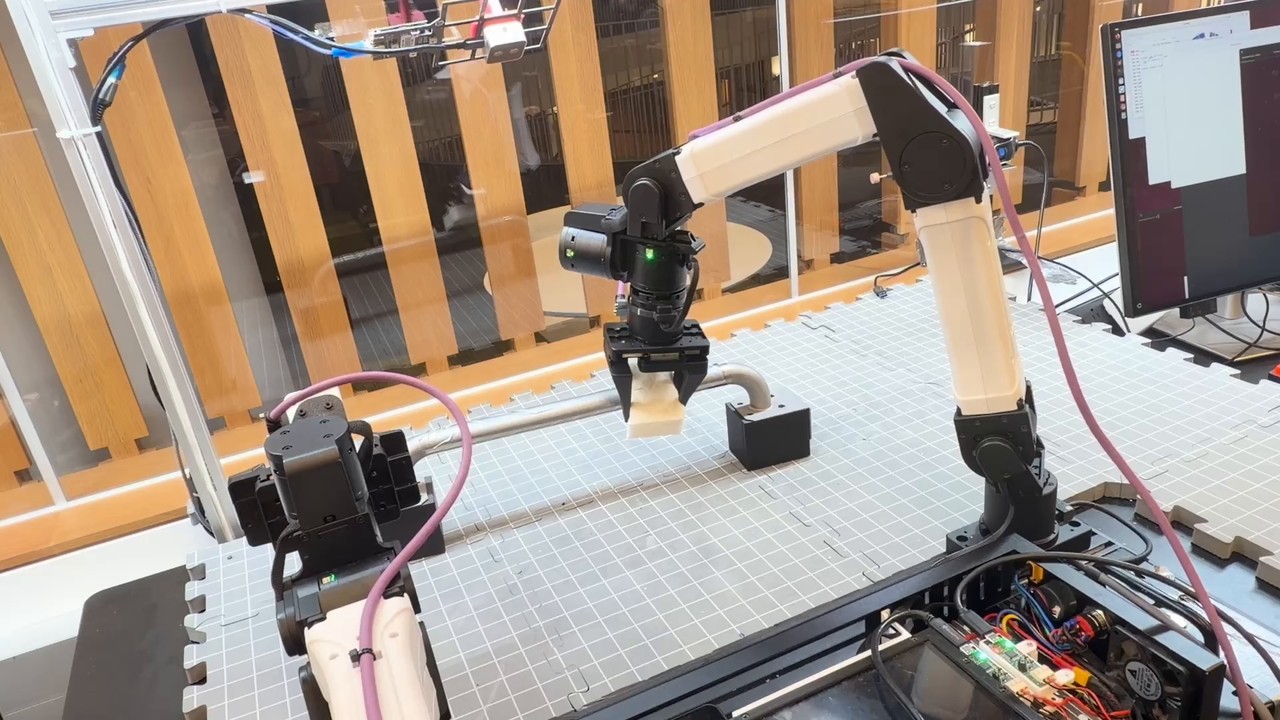}
    \caption{Sim2Real policy deployment.}
    \label{fig:r2s2r:real}
  \end{subfigure}\hfill
  \begin{subfigure}[t]{\rpw}
    \begin{tikzpicture}[inner sep=0pt, baseline=(p.south)]
      \node[anchor=south west] (p) at (0,0) {
        \includegraphics[width=\linewidth]{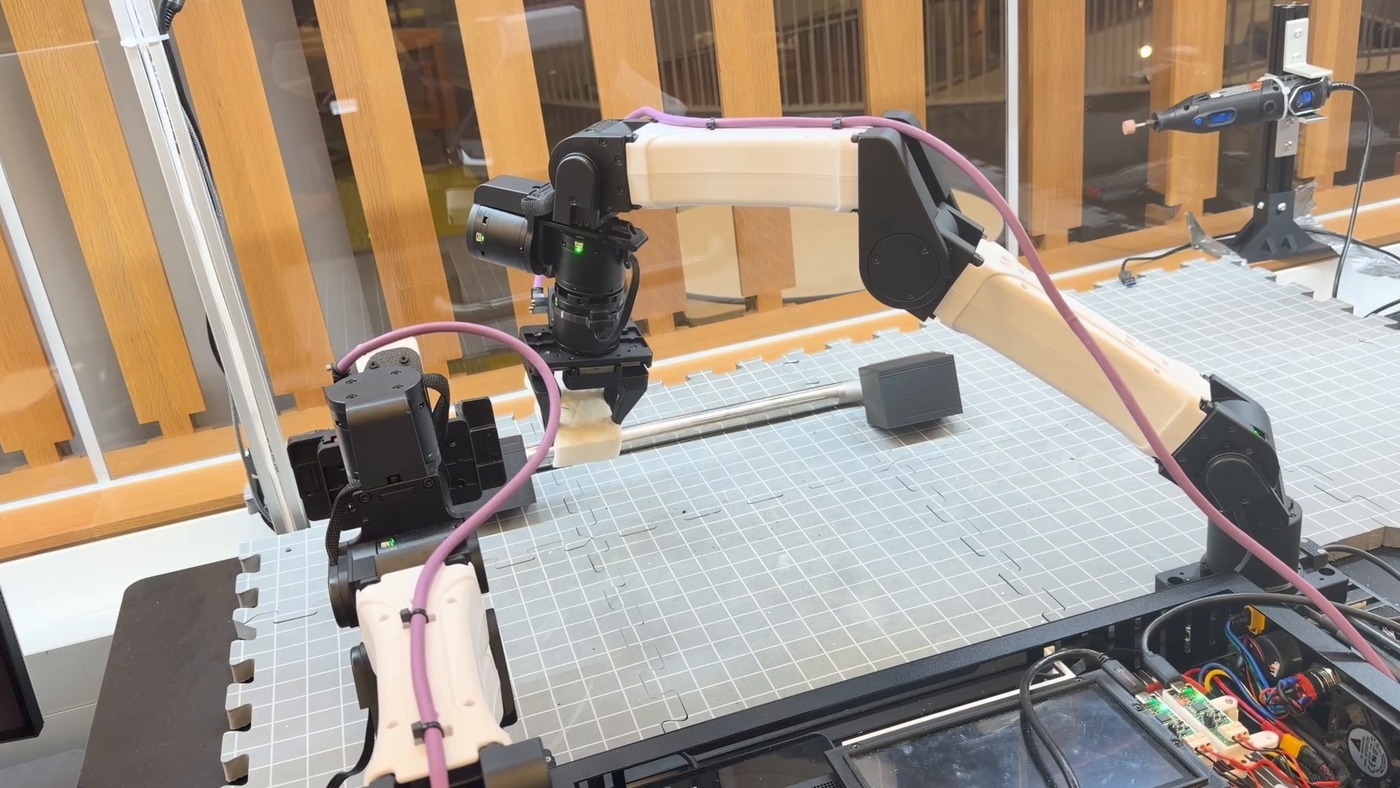}
      };
      \begin{scope}[x={(p.south east)}, y={(p.north west)}]
        \draw[red!85!black, line width=0.7pt,
          dash pattern=on 2pt off 1.2pt]
          (0.645,0.505) ellipse [x radius=0.075, y radius=0.11];
      \end{scope}
    \end{tikzpicture}
    \caption{Real-world failure from unmodeled physics.}
    \label{fig:r2s2r:fail}
  \end{subfigure}\hfill
  \begin{subfigure}[t]{\rpw}
    \tilestack{%
      \chiptile{\rlw}{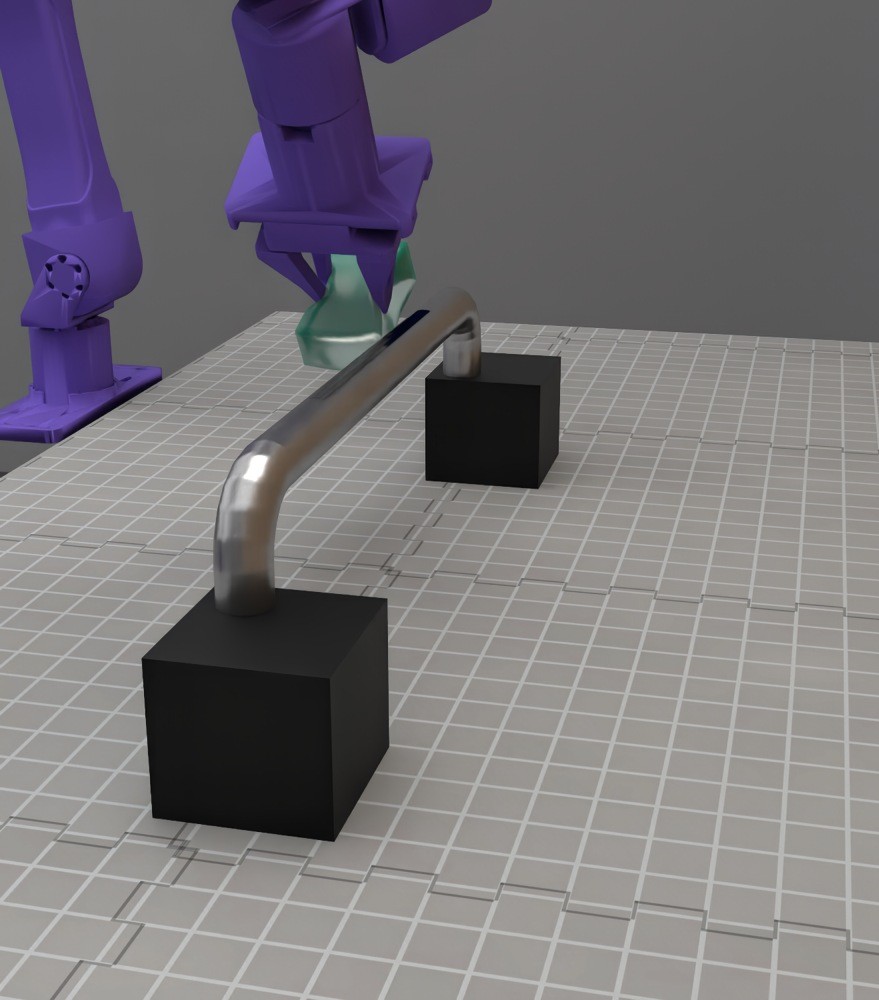}{original: fixed}\hfill
      \chiptile{\rlw}{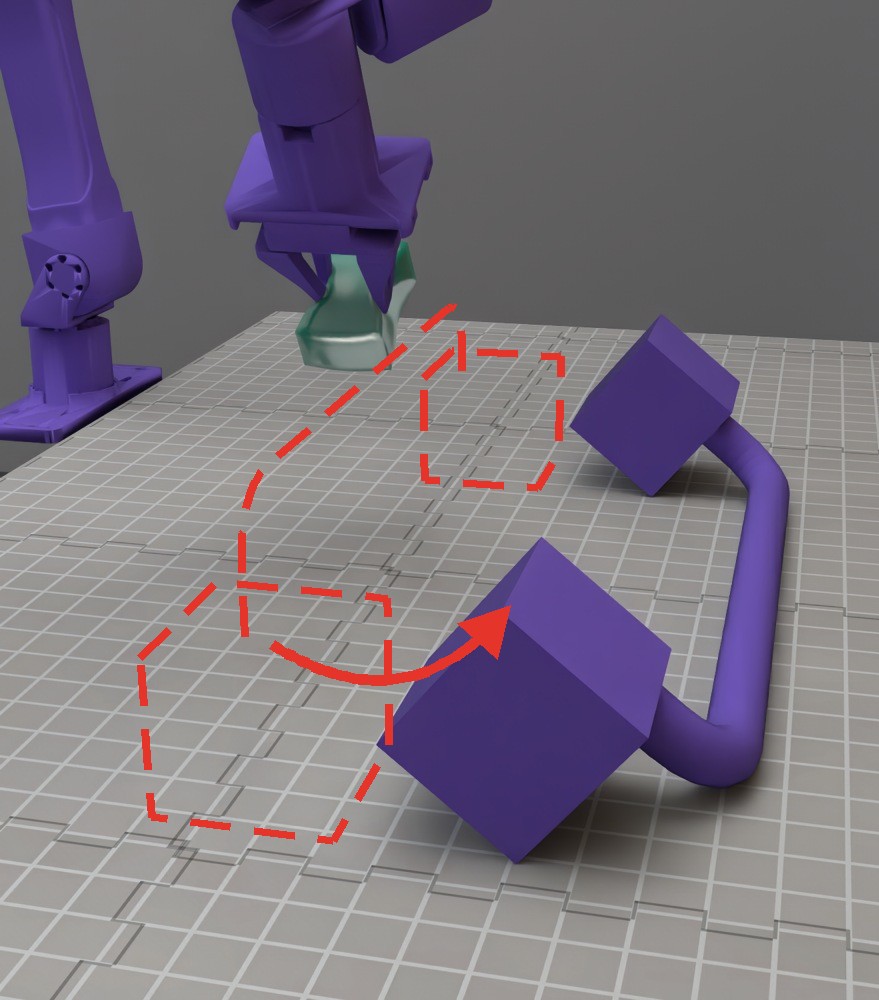}{revised: free}}
    \caption{Agentic simulation refinement.}
    \label{fig:r2s2r:revise}
  \end{subfigure}

  \caption{\textbf{Agentic Real2Sim2Real for sponge wiping.}
  (a)~A real-world bimanual manipulation task, where one arm holds a bar while the other wipes a marker stripe using a sponge.
  (b)~Simulation challenges involving robot motion, object-object contact, and robot-object contact.
  (c)~A coding agent constructs and iteratively refines the simulation through the interactive studio, completing a verified simulated wipe in 56 minutes.
  (d)~A multi-physics simulation in \rss.
  (e)~Physical variation of sponge stiffness and spatial variation of the bar and sponge across 30 generated layouts (bottom).
  (f)~GaP~\cite{gap2026} improves the simulated wiped area from 0\% to 100\% over 14 iterations using spatial-temporal simulation feedback from rss.
  (g)~The improved policy is deployed on the real robot to perform sponge wiping.
  (h)~A real-world failure caused by displacement of a bar support that was fixed in the original simulation.
  (i)~Given the observed failure, the coding agent revises the physical simulation to allow the bar and its supports to move under contact. Under an off-center wiping motion, the revised simulation captures bar displacement and tipping.}
  \label{fig:r2s2r}
\end{figure}
\subsection{Agentic Real2Sim2Real}
\label{sec:wipe}

We study agentic Real2Sim2Real using a bimanual sponge-wiping task, where one robot arm holds a bar while the other uses a foam sponge to wipe off a marker stripe (Fig.~\ref{fig:r2s2r:input}).
Given a smartphone video without CAD models or calibrated cameras, two off-the-shelf coding agents, Claude Fable~5.1 in Claude Code~\citep{claudecode2025} and GPT-6 Astra in Codex~\citep{openaicodex2025}, independently constructed multi-physics simulations in \rss in approximately one hour.
The agents estimated scene geometry from the video, configured robot and object assets, and iteratively refined the simulation using visual observations and spatial-temporal feedback (Fig.~\ref{fig:r2s2r:build}).
For example, simulator feedback helped identify errors in robot kinematics and contact conditions, allowing the agent to adjust the wiping motion and establish sponge contact with the bar.
One agent constructed and verified the simulated task within 56 minutes.

The resulting simulation couples MuJoCo-Warp within Newton for articulated and rigid bodies with SuperDex for sponge deformation, allowing contact forces and pressure distributions to be evaluated during wiping (Fig.~\ref{fig:r2s2r:twin}).
We then use \rss to vary the sponge's stiffness and generate 30 spatial layouts with different bar and sponge configurations (Fig.~\ref{fig:r2s2r:var}).
Using spatial-temporal simulation feedback, Graph-as-Policy (GaP)~\citep{gap2026} improved the simulated wiped area from 0\% to 100\% over 14 iterations (Fig.~\ref{fig:r2s2r:learn}).
The improved policy was subsequently deployed on the real robot to complete the wiping task (Fig.~\ref{fig:r2s2r:real}).
However, one real-world rollout revealed a failure in which the wiping motion displaced a bar support that was fixed in the simulation, causing the sponge to lose contact (Fig.~\ref{fig:r2s2r:fail}).
This failure highlights the importance of refining simulation models based on real-world feedback.

\textbf{Real-world Feedback and Simulation Refinement.}
We further study how \rss supports iterative simulation refinement using feedback from real-world deployment.
Given the failure video and a description of the observed failure, the coding agent identifies that the original simulation incorrectly fixes the bar supports to the environment.
The agent modifies the scene by replacing the fixed constraint with a free rigid body, allowing the bar and its supports to move under contact forces (Fig.~\ref{fig:r2s2r:revise}).
The agent then evaluates the revised simulation under different wiping trajectories and contact conditions.
While the original fixed-bar simulation cannot reproduce the observed displacement, the revised simulation captures bar motion and tipping when the sponge applies sufficient force or moves away from the bar's center.
This suggests how explicit physical constraints and editable scene parameters in \rss allow coding agents to identify missing physical interactions and refine the simulation based on real-world observations.
The revised simulation can then be used for further policy evaluation and improvement to reduce similar failures.

\section{Conclusion}
\label{sec:conclusion}

We present \rss, an interactive toolkit for simulating robot manipulation that makes the choice of solver, contact model, and physical parameters explicit and editable.
\rss supports multi-physics simulation by coupling solvers within Newton and external physics engines, and provides interactive interfaces for both humans and coding agents to construct and refine simulations.
We demonstrated its capabilities through 18 contact-rich manipulation tasks involving rigid bodies, articulated objects, cables, cloth, and soft-body deformation.
In a pilot study of agentic Real2Sim2Real, coding agents constructed a multi-physics simulation of a bimanual sponge-wiping task from a smartphone video in approximately one hour.
A Graph-as-Policy improved through simulation feedback was successfully deployed on the real robot, and real-world failure feedback was used to guide further simulation refinement.

\textbf{Limitations and Future Work.}
This paper reports a developing system with a pilot study suggesting the feasibility of interactive and agentic Real2Sim2Real.
The agentic simulation construction relies on existing material models, physics couplings, and robot assets provided by \rss, and its capabilities remain limited by the supported physics solvers.
Co-simulation with external physics engines also introduces computational overhead that may limit simulation throughput.
Furthermore, we evaluated sim-to-real transfer on a single manipulation task, and assessed the agreement between simulated and real-world behavior primarily through qualitative observations.
Future work will include more systematic quantitative evaluations across diverse manipulation tasks, including simulation accuracy, contact forces, object trajectories, agentic construction time, and real-world policy success rates.

\bibliography{refs}

\clearpage

\end{document}